\documentclass[letterpaper, 10 pt, conference]{ieeeconf}
\IEEEoverridecommandlockouts    
\usepackage{graphics}           
\usepackage{times}              
\usepackage{amsmath}            
\usepackage{amssymb}            
\usepackage{graphicx}
\usepackage{algorithm}
\usepackage[noend]{algpseudocode}
\usepackage{booktabs}
\usepackage{color}
\definecolor{instructioncolor}{rgb}{.5,.5,.5}

\usepackage[font=small]{caption}

\def\eqref#1{Eq.~(\ref{#1})}

\makeatletter
\usepackage{xspace}
\DeclareRobustCommand\onedot{\futurelet\@let@token\@onedot}
\def\@onedot{\ifx\@let@token.\else.\null\fi\xspace}

\makeatother

\usepackage{array}
\newcolumntype{L}[1]{>{\raggedright\let\newline\\\arraybackslash\hspace{0pt}}m{#1}}
\newcolumntype{C}[1]{>{\centering\let\newline\\\arraybackslash\hspace{0pt}}m{#1}}
\newcolumntype{R}[1]{>{\raggedleft\let\newline\\\arraybackslash\hspace{0pt}}m{#1}}

\usepackage[utf8]{inputenc}
\usepackage[T1]{fontenc}

\usepackage{hyperref}   
\usepackage{cleveref}   
\usepackage{xcolor}     
\usepackage{multirow}
\usepackage{makecell}
\usepackage{xcolor} 
\usepackage{textcomp} 
\definecolor{negativecolor}{RGB}{0, 163, 226} 
\definecolor{positivecolor}{RGB}{173, 69, 31} 

\hypersetup{
    colorlinks=true,
    linkcolor=blue,
    citecolor=blue,
    filecolor=blue,
    urlcolor=blue,
}

\crefformat{figure}{Fig.~#2\textcolor{red}{#1}#3}

\crefformat{table}{Table~#2\textcolor{red}{#1}#3}

\title{\LARGE \bf LuSeg: Efficient Negative and Positive Obstacles Segmentation via Contrast-Driven Multi-Modal Feature Fusion on the Lunar}

\author{Shuaifeng Jiao \and Zhiwen Zeng \and Zhuoqun Su \and Xieyuanli Chen \and Zongtan Zhou$^*$ \and Huimin Lu 
  \thanks{All authors are with the College of Intelligence Science and Technology, and the National Key Laboratory of Equipment State Sensing and Smart Support, National University of Defense Technology.}%
  \thanks{Corresponding author: Zongtan Zhou (narcz@nudt.edu.cn).}
  \thanks{This work was supported in part by the National Science Foundation of China (Grant No. 62403478 and 62203460), Young Elite Scientists Sponsorship Program by CAST (No. 2023QNRC001).
  }%
}

\begin{document}
\maketitle
\thispagestyle{empty}
\pagestyle{empty}

\begin{abstract}As lunar exploration missions grow increasingly complex, ensuring safe and autonomous rover-based surface exploration has become one of the key challenges in lunar exploration tasks. In this work, we have developed a lunar surface simulation system called the Lunar Exploration Simulator System (LESS) and the LunarSeg dataset, which provides RGB-D data for lunar obstacle segmentation that includes both positive and negative obstacles. Additionally, we propose a novel two-stage segmentation network called LuSeg. Through contrastive learning, it enforces semantic consistency between the RGB encoder from Stage I and the depth encoder from Stage II. Experimental results on our proposed LunarSeg dataset and additional public real-world NPO road obstacle dataset demonstrate that LuSeg achieves state-of-the-art segmentation performance for both positive and negative obstacles while maintaining a high inference speed of approximately 57\,Hz. We have released the implementation of our LESS system, LunarSeg dataset, and the code of LuSeg at: \href{https://github.com/nubot-nudt/LuSeg.git}{https://github.com/nubot-nudt/LuSeg}.
\end{abstract}

\section{Introduction}
\label{sec:intro}
The Lunar, being the closest celestial body to Earth, has garnered an unprecedented level of interest in exploration like never before. With the increasing complexity of lunar exploration missions, the lengthy Earth-Moon communication link restricts real-time human operation and control of lunar rovers~\cite{frank2016AI}. Consequently, there is a growing demand for autonomy in unmanned systems such as lunar rovers. Achieving autonomous, safe, and efficient lunar surface exploration with lunar rovers is one of the critical challenges that cannot be overlooked in lunar exploration missions. Fast and accurate obstacle segmentation is one of the key techniques for ensuring the safe and efficient exploration of the lunar surface by rovers. Obstacles can be broadly classified into two types: positive and negative. Positive obstacles, such as lunar rocks, obstruct the rover's path, while negative obstacles, such as craters, pose a risk of entrapment, potentially compromising the success of explorations.

The development and validation of perception algorithms for lunar rovers require extensive datasets that encompass a variety of terrains and environmental conditions. Due to the inherent limitations in data collection scenarios during actual lunar exploration missions, along with the challenges and high costs associated with constructing diverse lunar analog terrains on Earth, the development of a lunar surface simulation system becomes particularly important. Existing lunar surface simulation systems are often not publicly accessible~\cite{crues2022dles, bingham2023dust, Allan2019PlanetaryRS}, or suffer from limited functionality and low rendering quality~\cite{martinez2023multi, orsula2022iros}, highlighting the urgent need for a high-quality, user-friendly, and flexible robotic lunar simulation system. In this work, we have developed a Lunar Exploration Simulator System (LESS) based on Unreal Engine 4 (UE4) and AirSim~\cite{shah2018airsim}. LESS supports the Robot Operating System (ROS) and commonly used robotic sensors, while also allowing for scene and environment customization. Researchers can integrate additional sensors or customize terrain models for specialized applications.
\begin{figure}[t]
  \centering
  \includegraphics[width=0.95\linewidth]{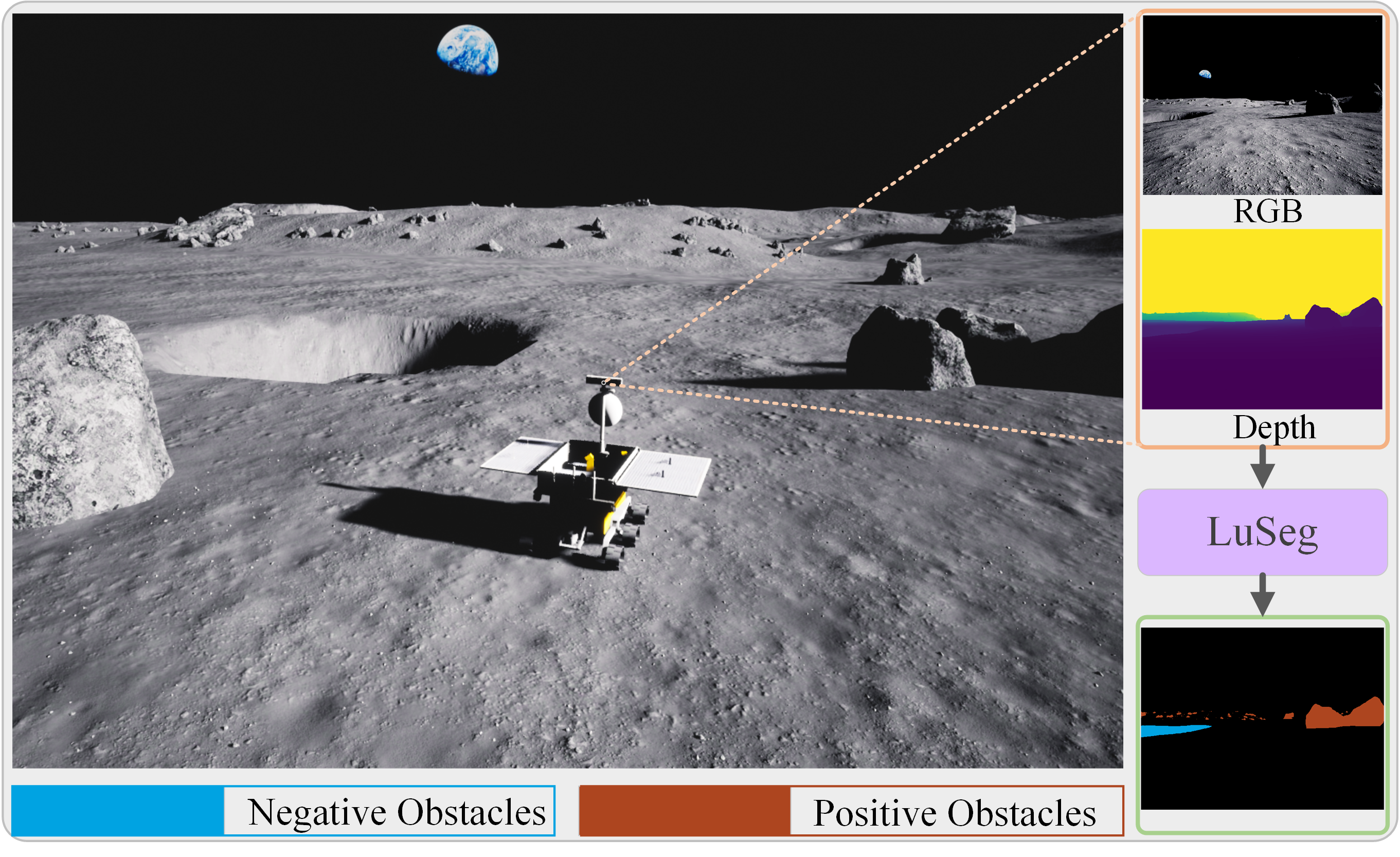}
  \caption{The scene of a lunar rover driving safely and autonomously in the Lunar Exploration Simulator System }
  \label{fig:motivation}
\end{figure}

To tackle obstacle segmentation, deep learning algorithms demonstrate promising performance while requiring labeled datasets for training and testing models. Current lunar semantic segmentation datasets are often limited in scope, typically containing only positive obstacles such as lunar rocks~\cite{pessia2020, boerdijk2023resyris}. Based on our LESS system, we have constructed the LunarSeg dataset, which encompasses a variety of terrains and lighting conditions, while also including both positive and negative obstacles. To our knowledge, there are very few publicly available datasets for lunar obstacle semantic segmentation, while our proposed LunarSeg includes both negative obstacles and positive obstacles. 
Several single-modal segmentation methods (e.g., RGB) and corresponding datasets have been proposed for detecting positive and negative obstacles in typical automotive driving scenarios~\cite{muhammad2022tits,ma2022computer}. However, single-modal semantic segmentation suffers from information deficiency and limited robustness, particularly under extreme lighting conditions and complex environmental variations, which are common on the lunar surface. Many works~\cite{seichter2021icra,zhou2023tase} have demonstrated that the fusion of depth images with RGB images can significantly enhance performance in semantic segmentation. Early fusion methods~\cite{couprie2013iclr} necessitate precise spatial or temporal alignment, which is often difficult to achieve in practical applications. Late fusion methods~\cite{valada2017deep,cheng2017cvpr} may not adequately capture fine-grained relationships between modalities, as the fusion occurs at a higher level of features. Although intermediate fusion aims~\cite{zhou2023tiv,du2024cvpr} to integrate semantic information from different modalities, it may encounter challenges related to poor alignment, and the incorporation of intermediate fusion modules can lead to increased computational costs and complexity. To address these challenges, we propose a novel two-stage training segmentation method that effectively maintains the semantic consistency of multimodal features through our proposed Contrast-Driven Fusion module (CDFM). Notably, the CDFM module is not utilized during the inference phase, ensuring that no additional computational overhead is introduced. Extensive experiments conducted on the public NPO dataset and our LunarSeg dataset demonstrate that our approach outperforms baseline methods in both positive and negative obstacle segmentation performance.

In summary, the contributions of this work are threefold:
\begin{itemize}
    \item We have developed and publicly released a lunar simulation system named LESS, along with the LunarSeg dataset, a novel lunar dataset specifically designed for obstacle segmentation that includes both positive and negative obstacles.
    \item We propose a novel two-stage training segmentation method that effectively maintains the semantic consistency of multimodal features via our proposed Contrast-Driven Fusion module.
    \item Extensive experimental results on our LunarSeg and additional public NPO datasets show that our proposed method can achieve state-of-the-art performance.
\end{itemize}

\section{Related Work}
\label{sec:related}
\subsection{Lunar Simulator System}
Various lunar simulator systems have been developed for lunar exploration. Orsula et al.~\cite{orsula2022iros} conducted research on robotic grasping strategies within a lunar simulation environment developed using Gazebo. However, it lacks terrain environments suitable for other vision-based tasks. DLES~\cite{crues2022dles} utilizes digital elevation models (DEM) from LRO and LOLA to accurately simulate the lunar south pole. In addition to terrain generation, it incorporates features not captured in the DEM, such as small craters and rocks of various sizes, to more realistically replicate the lunar surface. DUST~\cite{bingham2023dust} incorporates the enhanced terrain from the DLE product to facilitate the exploration of the lunar south pole and its intricate illumination conditions. It achieves highly accurate rendering of surface features and terrain by utilizing Unreal Engine 5's double-precision positioning, multi-source dynamic lighting, and detail optimization techniques. Unfortunately, DUST and DLES don't support robotic systems and are not available to the public.  NASA~\cite{Allan2019PlanetaryRS} has made extensive modifications to the Gazebo rendering engine, developing a lunar exploration simulator that features capabilities such as lunar rover navigation, system monitoring, and scientific instrument simulation. Regrettably, this system is also not publicly accessible.

\subsection{Single-Modal Semantic Segmentation Networks}
Researchers have proposed and developed a wide array of efficient single-modal semantic segmentation algorithms grounded in deep learning. DAFormer~\cite{hoyer2022cvpr} leverages a Transformer encoder alongside a multilevel context-aware feature fusion decoder to augment feature representation and semantic comprehension. LAANet~\cite{zhang2022laanet} incorporates an efficient asymmetric bottleneck module and an attention-guided multi-scale context information aggregation module, facilitating high accuracy in real-time semantic segmentation. PIDNet~\cite{xu2023cvpr} integrates convolutional neural networks with proportional-integral-derivative (PID) controllers, utilizing boundary attention to effectively guide the fusion of detailed and contextual information. Nevertheless, single-modal semantic segmentation struggles to adequately address the challenges of information scarcity and diminished robustness, particularly in the presence of extreme lighting conditions and environmental complexities, such as those encountered on the lunar surface.

\begin{figure*}[t]
  \centering
  \includegraphics[width=1\linewidth]{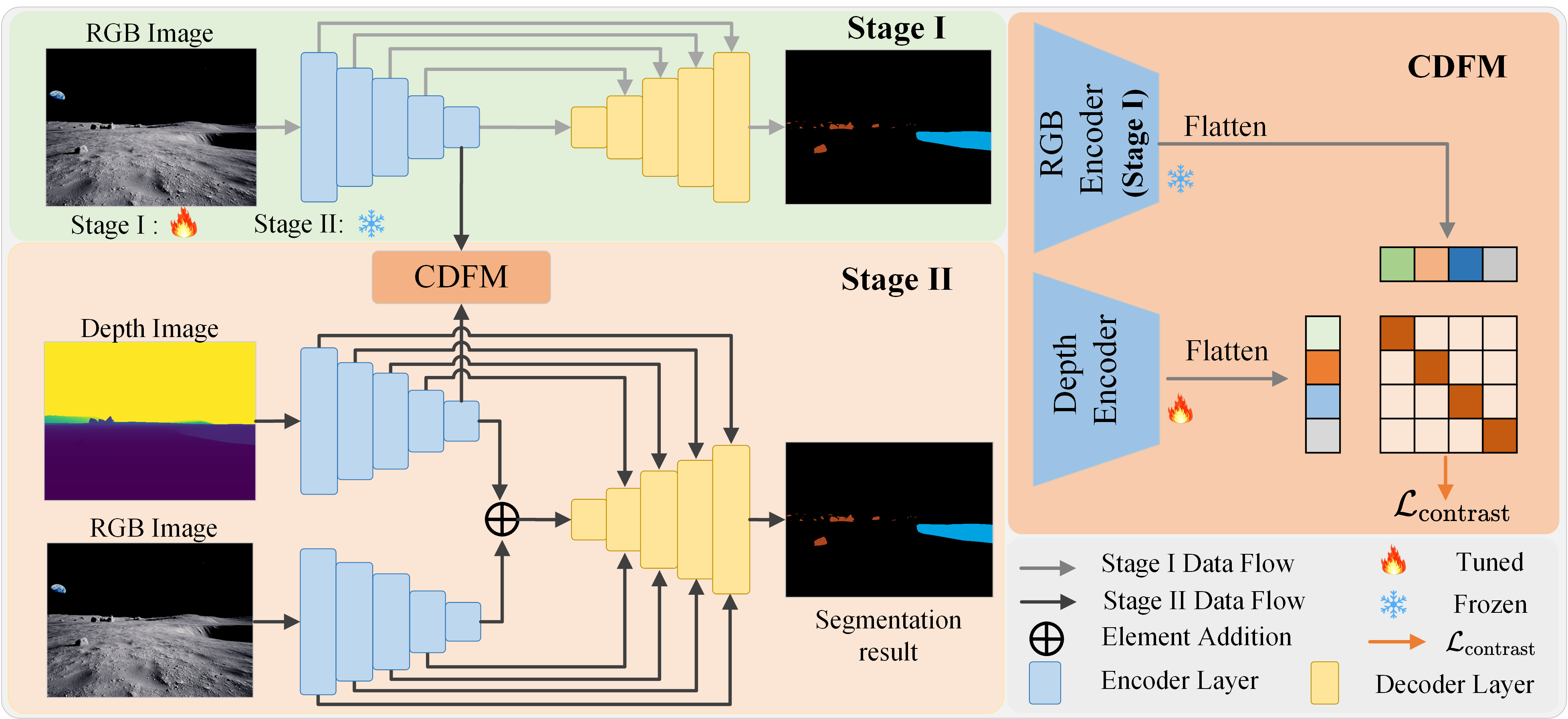}
  \caption{Overall architecture of our proposed LuSeg. There are two stages: Stage I involves single-modal training using only RGB images as input, while Stage II focuses on multi-modal training with both RGB and depth images as input. In Stage II, the output of the depth encoder is aligned with the output of the RGB encoder from Stage I, whose parameters are frozen during this stage. This serves as input to our proposed Contrast-Driven Fusion Module (CDFM). The final output of Stage II is the result of our LuSeg.}
  \label{fig:framework}
\end{figure*}

\subsection{Multi-Modal Semantic Segmentation Networks}
Early investigations into multimodal semantic segmentation concentrated on two principal fusion strategies: early fusion and late fusion~\cite{qian2021tits}. Early fusion techniques directly integrate the RGB and depth channels within an unimodal network~\cite{couprie2013iclr}, necessitating precise spatial or temporal alignment, which is often difficult to fulfill in practice. In contrast, late fusion methods combine modality-specific features following independent processing by separate encoders, employing techniques such as concatenation or element-wise summation for integration. The LSD-GF~\cite{cheng2017cvpr} framework introduced a gated fusion layer designed to automatically amalgamate high-level features from both modalities, thereby facilitating effective information fusion. However, this higher-level fusion may inadequately capture the fine-grained intermodal relationships.

In recent years, significant efforts have been dedicated to the development of advanced fusion modules aimed at facilitating hierarchical or intermediate fusion. CACFNet~\cite{zhou2023tiv} introduces a cross-modal attention fusion module to fuse the complementary information from RGB and thermal images. AsymFormer~\cite{du2024cvpr} presents the LAFS and CMA modules to enable selective multimodal feature fusion across spatial and channel dimensions. BCINet~\cite{zhou2023if} leverages BCIM to enhance single-modal features via FEM and integrates them across sides to capture complementary cross-modal information. SGACNet~\cite{zhang2023sj}  enhances RGB-D feature extraction by using a channel and spatial attention fusion module in the encoder to selectively merge depth features with RGB features, ensuring robustness while reducing computational parameters. Due to semantic inconsistency between different modalities when integrating intermediate features, such methods may not fully exploit the complementary nature of modalities. In contrast, our proposed approach employs contrastive learning to facilitate feature alignment, enabling the semantic consistency of multimodal features and significantly improving segmentation performance.

\subsection{Datasets for Extraterrestrial Segmentation }

Artificial Lunar Landscape Dataset~\cite{pessia2020}  is derived from a collection of lunar images simulated using Terragen software by Planetside. It currently comprises 9,766 high-fidelity rendered images depicting lunar rocky landscapes, each paired with corresponding semantic segmentation labels. These labels include regional annotations for categories such as sky, small rocks, and large rocks. The ReSyRIS~\cite{boerdijk2023resyris} dataset is specifically designed for rock instance segmentation, containing both real images captured in the lunar simulation environment at Mt. Etna, annotated with pixel-level instance masks, as well as synthetic images generated by the Blender-based OAISYS simulator~\cite{muller2021iros}. The AI4Mars dataset~\cite{swan2021cvpr} consists of 35,000 images captured by the Curiosity, Opportunity, and Spirit rovers, covering four primary categories: soil, bedrock, sand, and boulders, with nearly 326,000 semantic segmentation labels. Particularly, these datasets focus solely on positive obstacles such as rocks, and exclude negative obstacles like craters.

\section{Our LuSeg Obstacle Segmentation Network}
\label{sec:main}

\subsection{The Overall Architecture}
To tackle the challenge of feature distribution discrepancies between RGB and depth images in multimodal segmentation tasks, we introduce a multimodal segmentation network based on a contrast-driven fusion strategy, named LuSeg. LuSeg employs a two-stage training framework that effectively ensures semantic consistency of multimodal features, significantly enhancing segmentation performance. 

The overall architecture of our proposed LuSeg is illustrated in~\cref{fig:framework}. In Stage I, the framework comprises a five-layer encoder and a five-layer decoder, which are interconnected through skip connections between corresponding layers. We utilize ResNet-152~\cite{he2016cvpr} as the encoder to proficiently extract features from the input data. The decoder is adapted from InconSeg~\cite{feng2023ral}. Within the encoder, each layer systematically reduces the resolution of the input image by half. In the decoder, each layer restores the resolution, while simultaneously halving the number of channels in the feature map. Stage I encompasses single-modal training, employing only RGB images as input. Upon completion of this training phase, the weights of the RGB encoder are frozen in Stage II, thereby providing pre-trained RGB features that serve as references for depth modality alignment within our proposed Contrast-Driven Fusion Module (CDFM).

\begin{figure}[t]
  \centering
  \includegraphics[width=1\linewidth]{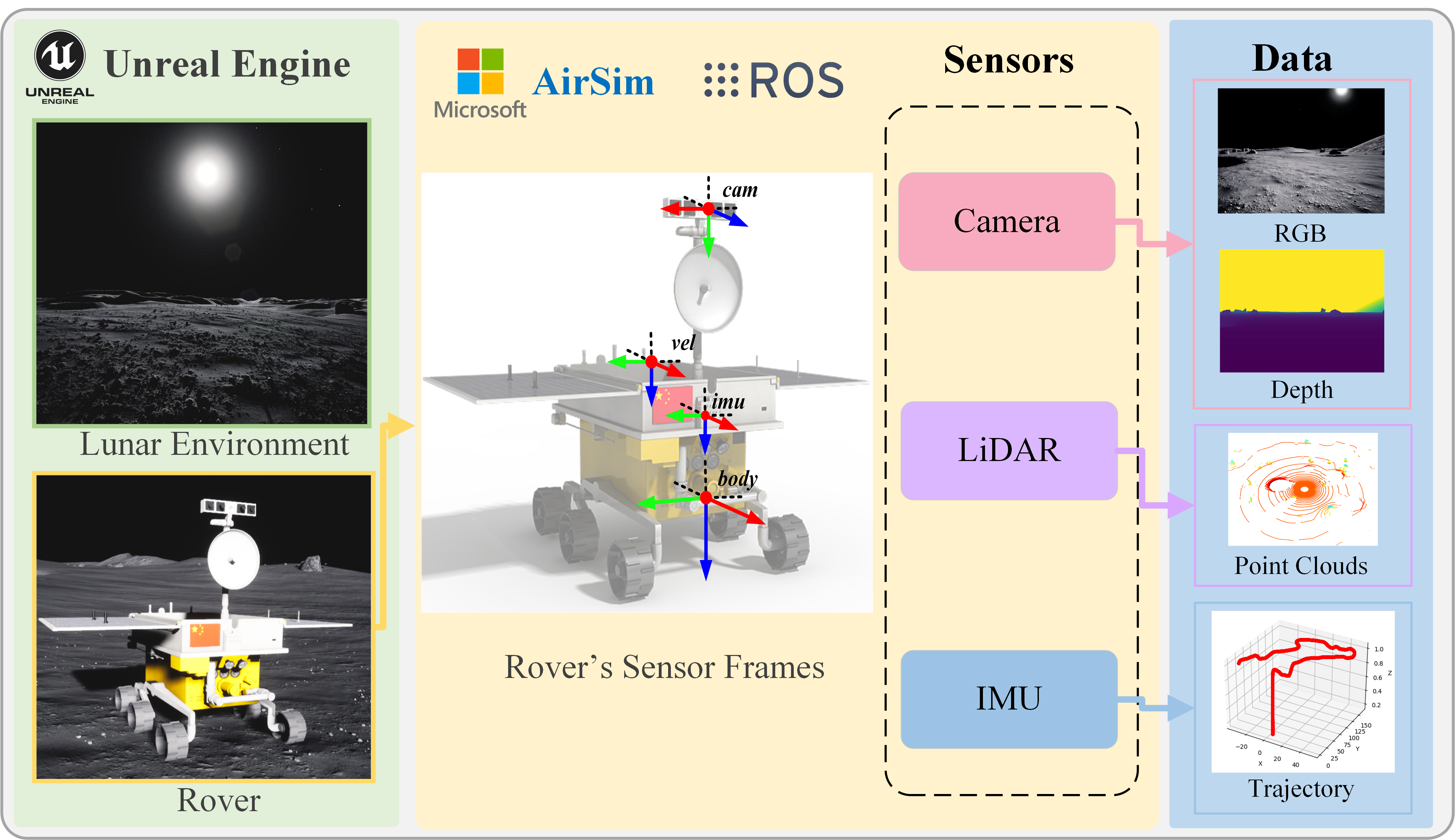}
  \caption{The Overall of Lunar Exploration Simulator System(LESS).}
  \label{fig:less}
\end{figure}

Stage II is focused on multi-modal training, utilizing both RGB and depth images as inputs. In this architecture, the depth and RGB encoders, along with the decoder, preserve the structural framework from Stage I, with both encoders connected to corresponding layers of a shared decoder through skip connections. The outputs from the depth encoder in Stage II and the frozen RGB encoder from Stage I are subsequently integrated into the Contrast-Driven Fusion Module (CDFM), enabling the creation of a multi-modal aligned representation that exhibits enhanced semantic consistency and enriched geometric features. These aligned features mainly focus on improving semantic coherence and geometric fidelity. However, this emphasis may unintentionally obscure certain unique aspects of the RGB modality, such as color richness and textural detail. To mitigate this limitation, the RGB encoder in Stage II contributes task-specific, fine-grained RGB features, which are seamlessly integrated with the depth encoder’s aligned outputs through the CDFM module, thereby fully leveraging multi-modal information.

\subsection{The Contrast-Driven Fusion Module}
As aforementioned, there are inherent differences in the spatial and geometric information representation between RGB images and depth images, leading to inconsistencies in the semantic features of the two modalities, which in turn affects the effective utilization of the information from both modalities. To tackle this issue, we propose a Contrastive Driven Fusion Module (CDFM). This module aligns the features of RGB and depth modalities through a contrastive learning mechanism to ensure semantic consistency between the different modalities.

The structure of our CDFM module is shown on the right-hand side of ~\cref{fig:framework}. The module has two inputs: the output of the depth encoder in Stage II and the output of the frozen RGB encoder from Stage I. The feature maps output by the depth encoder(Stage II) and the RGB encoder(Stage I) are flattened and denoted as $\boldsymbol{f}_{depth} \in \mathbb{R}^{N\times D}$and $\boldsymbol{f}_{RGB} \in \mathbb{R}^{N\times D}$, respectively. We employ a contrastive learning procedure~\cite{chen2020icml} to bridge the modality gap between depth and RGB features, enabling precise alignment of semantic features as:
\begin{equation}
\begin{aligned}
&\text{sim}(\boldsymbol{f}_{{depth}_i}, \boldsymbol{f}_{{RGB}_j}) = \frac{\boldsymbol{f}_{{depth}_i} \cdot \boldsymbol{f}_{{RGB}_j}}{\|\boldsymbol{f}_{{depth}_i}\| \|\boldsymbol{f}_{{RGB}_j}\|}, \\
&\mathcal{L}_{cont} = - \frac{1}{N}\sum_{i=1}^{N} \log\frac{\exp(\text{sim}(\boldsymbol{f}_{{depth}_i}, \boldsymbol{f}_{{RGB}_i}) / \tau)}
{\sum_{k=1}^{N} \exp(\text{sim}(\boldsymbol{f}_{{depth}_i}, \boldsymbol{f}_{{RGB}_k}) / \tau)}
\label{eq:NT-Xent}
\end{aligned}
\end{equation}
where $\text{sim}(\cdot)$ denotes the computation of cosine similarity, $N$ represents the batch size, and $\tau$ is the temperature parameter utilized to modulate the sharpness of the similarity distribution, with a specified value of 0.5 for this study.

The principal objective of this loss function is to facilitate the clustering of features corresponding to the same semantic within the embedding space, while concurrently maintaining distinctiveness among features of disparate semantic or non-corresponding features, thus promoting semantic consistency across modalities. Importantly, the CDFM module is not involved during the inference phase, ensuring that no additional computational overhead is introduced.

\begin{figure}[t]
  \centering
  \includegraphics[width=1\linewidth]{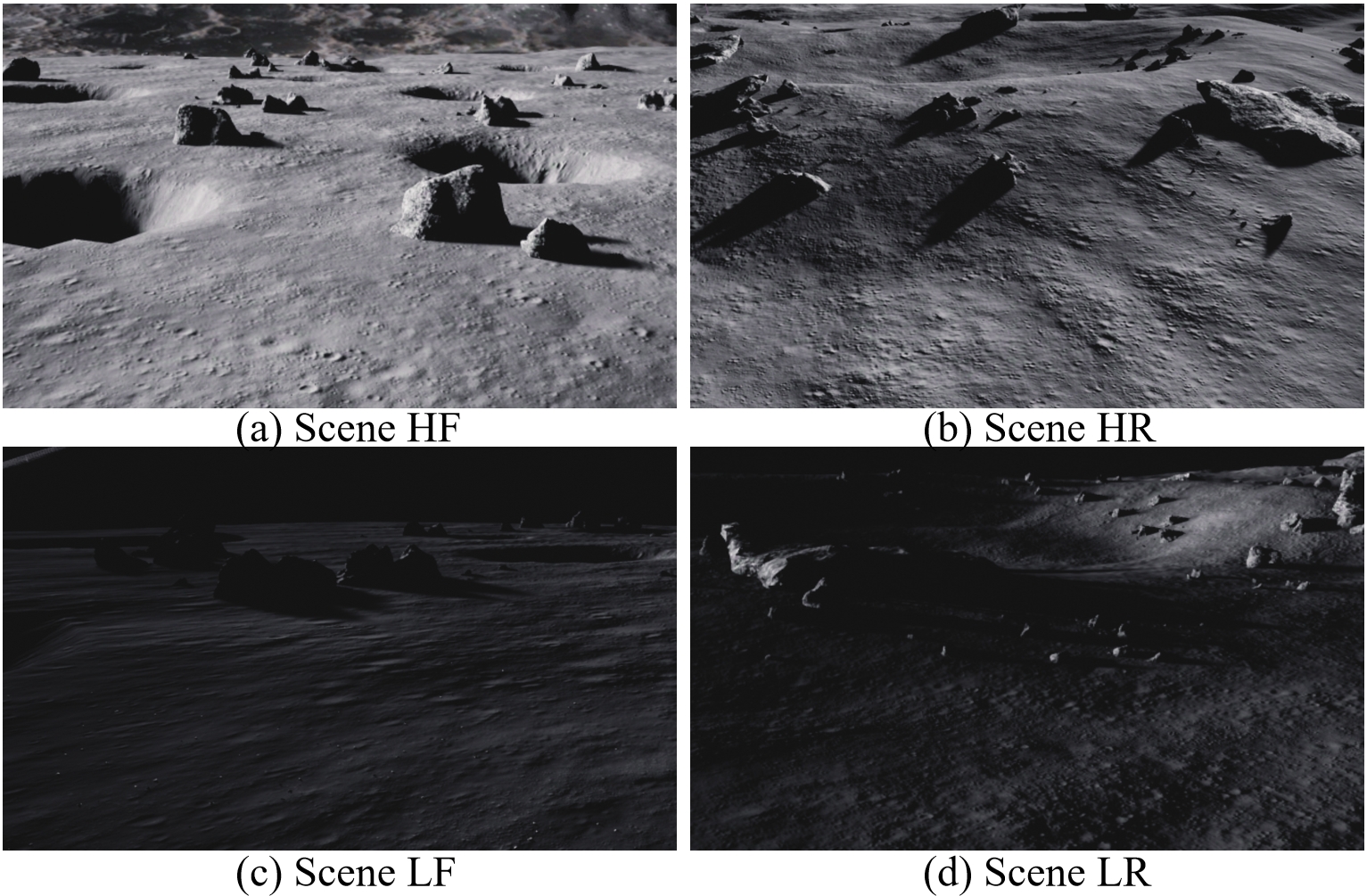}
  \caption{Various lunar surface scenes contained in the LunarSeg dataset. (a) high illumination and flat terrain (HF), (b) high illumination and rough terrain (HR), (c) low illumination and flat terrain (LF), and (d) low illumination and rough terrain (LR).}
  \label{fig:scenes}
\end{figure}

\subsection{The Loss Functions}
For our multimodal image segmentation tasks, we here also incorporate the Lov\'{a}sz-Softmax~\cite{berman2018cvpr} to directly optimize the Intersection-over-Union (IoU) metric. While IoU is widely used to evaluate segmentation accuracy by measuring the overlap between predictions and ground truth, it is non-differentiable and cannot be directly integrated into gradient-based training. To address this,~\cite{berman2018cvpr} leverages the Lovász extension, which relaxes discrete IoU into a continuous surrogate by embedding it into a hypercube space. This relaxation enables gradient computation for end-to-end optimization. The Lov\'{a}sz-Softmax loss (\(\mathcal{L}_{ls}\)) is defined as:
\begin{equation}
\begin{aligned}
\mathcal{L}_{ls} &= \frac{1}{|C|}\sum_{c\in C}\overline{\Delta_{J_c}}(m(c)), \quad \text{and} \\
m_i(c) &= \begin{cases} 
1 - x_i(c) & \text{if } c = y_i(c) \\
x_i(c) & \text{otherwise},
\end{cases}
\end{aligned}
\end{equation}
where \(|C|\) represents the class number, \(\overline{\Delta_{J_c}}\) defines the Lovász extension of the Jaccard index, \( x_i(c) \in [0,1] \) and \( y_i(c) \in [-1,1] \) hold the predicted probability and ground truth label of pixel \(i\) for class \(c\), respectively.

Finally, the total loss function of LuSeg is a linear combination of both contrast  and \textit{Lov\'{a}sz-Softmax} losses as follows:

\begin{equation}
\begin{aligned}
\mathcal{L} &= \mathcal{L}_{cont} + \mathcal{L}_{ls}
\end{aligned}
\label{eq:loss}
\end{equation}

\begin{figure}[t]
  \centering
  \includegraphics[width=1\linewidth]{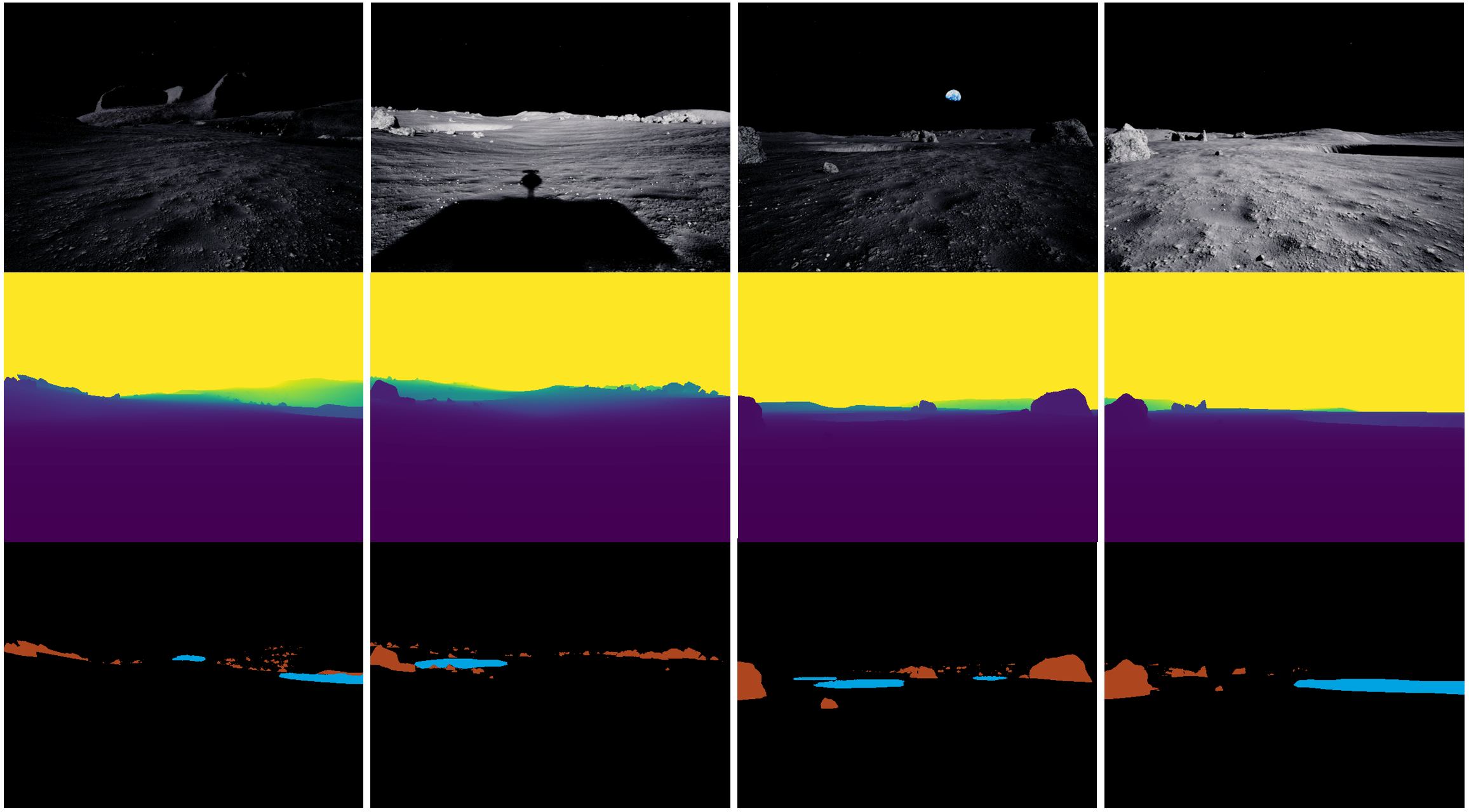}
  \caption{Sample RGB images, depth images, and ground truth in our LunarSeg dataset. \textcolor{negativecolor}{\rule{1.5ex}{1.5ex}} and \textcolor{positivecolor}{\rule{1.5ex}{1.5ex}} represent negative obstacles and positive obstacles, respectively.}
  \label{fig:dataset_sample}
\end{figure}

\section{Our Lunar Obstacle Segmentation Dataset}
\label{sec:data}
\subsection{Lunar Exploration Simulator System and Data Collection}
In this work, we build and release a lunar surface simulation system based on Unreal Engine and AirSim~\cite{shah2018airsim}, called the Lunar Exploration Simulator System (LESS), along with a large-scale dataset that contains both positive and negative obstacles for lunar obstacle segmentation.

As illustrated in~\cref{fig:less},  the LESS system integrates a high-fidelity lunar terrain model, a customizable rover platform, and a multi-modal sensor suite, while also supporting the Robot Operating System (ROS) to enable realistic data generation and the validation of autonomous perception algorithms for the rover. The system utilizes the advanced rendering capabilities of Unreal Engine to simulate lunar terrain and dynamic lighting conditions, including intense shadows and extreme brightness variations.  The rover is equipped with various adjustable sensors, including RGB cameras, LiDAR, and an Inertial Measurement Unit (IMU), which interact with the simulated lunar environment to collect synchronized data streams. Camera sensors capture high-resolution RGB imagery at adjustable frame rates, while LiDAR generates dense 3D point clouds with configurable angular resolution and range parameters. Sensor frames are spatially and temporally aligned to ensure consistency across modalities, and trajectory data (e.g., rover pose, velocity) are recorded to support downstream tasks like SLAM and motion planning. LESS provides a scalable platform for developing and validating perception algorithms in extraterrestrial environments. This open-source framework is designed for high extensibility, allowing researchers to integrate additional sensors or customize terrain models according to the specific requirements of their applications.

The dataset is collected using an RGB-D camera mounted on the rover within the LESS system. The RGB-D sensor operated at a resolution of 640×480 pixels and a frame rate of 10 Hz, capturing synchronized RGB images and 16-bit depth maps. To maximize environmental diversity, we procedurally generated four distinct lunar scenarios as shown in~\cref{fig:scenes}, systematically varying the intensity of the illumination (high/low) and the complexity of the terrain (flat/rough). We collect a total of 5,143 image pairs across the four scenarios, with each pair consisting of an RGB image and a 16-bit depth map. 

We name our dataset the LunarSeg dataset, which, to the best of our knowledge, is one of the few publicly available datasets specifically developed for lunar obstacle segmentation. Unlike existing datasets, LunarSeg includes both positive and negative obstacles, addressing a critical gap in this field. By open-sourcing this dataset, we aim to facilitate future research in lunar terrain understanding and autonomous navigation. Sample images from the dataset are presented in~\cref{fig:dataset_sample}.

\subsection{Dataset Analysis}
\begin{figure}[t]
  \centering
  \includegraphics[width=1\linewidth]{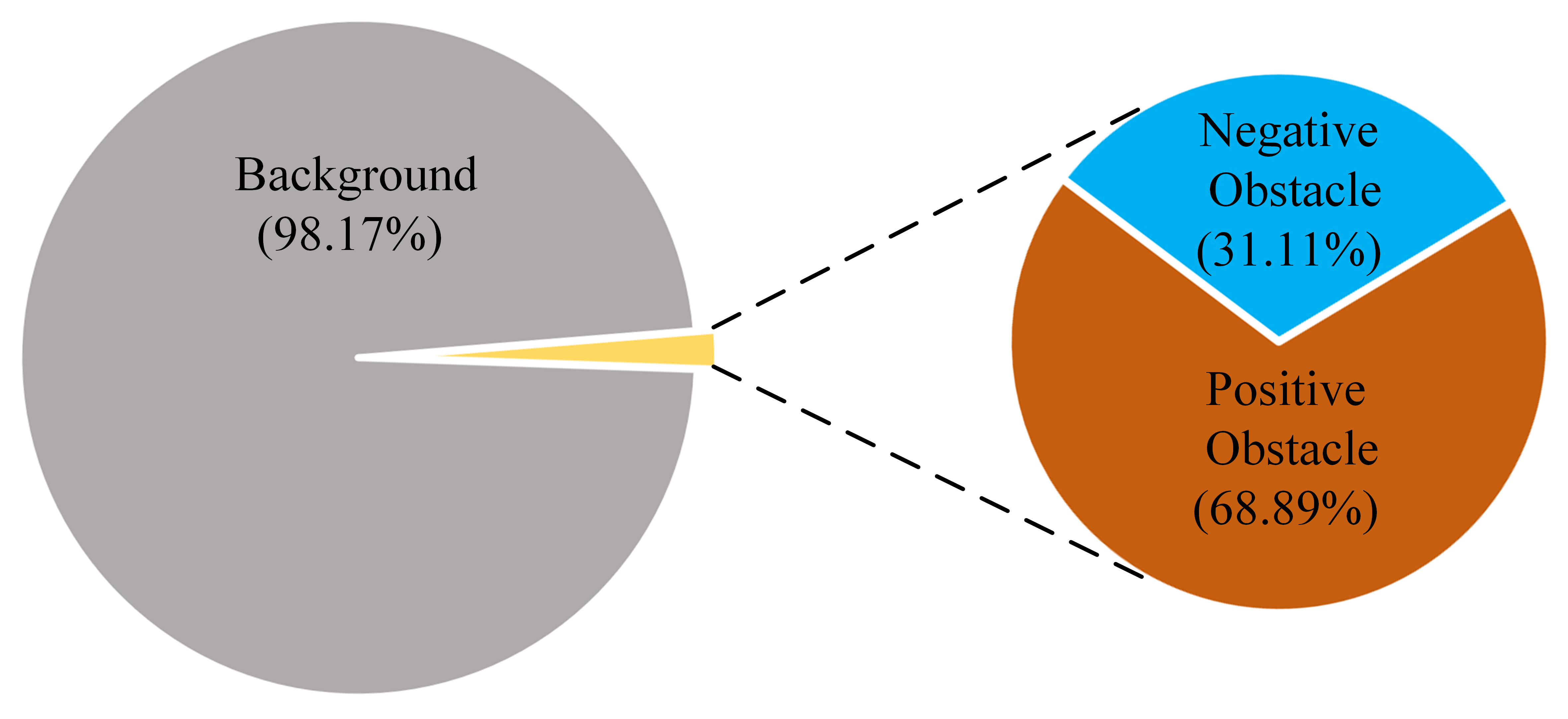}
  \caption{Pixel ratio for each class in our LunarSeg dataset.}
  \label{fig:pixs_ratio}
\end{figure}

All image pairs were manually annotated by a professional team to label negative obstacles (e.g., craters) and positive obstacles (e.g., lunar rocks). The LunarSeg dataset comprises 1650 images captured from low illumination and rough terrain (LR) scenes, 1270 images from low illumination and flat terrain (LF) scenes, 1144 images from high illumination and tough terrain (HR) scenes, and 812 images from high illumination and flat terrain (HF) scenes. Within the LunarSeg dataset, 4061 images include negative obstacles and 5143 images include positive obstacles. ~\cref{fig:pixs_ratio} shows the pixel ratio for each class.

\section{Experimental Evaluation}
\label{sec:exp}

\begin{table*}[t]
  \caption{The Comparetive Result(\%) on the Testing Set of   Our LunarSeg Dataset.}
  \centering
\renewcommand\arraystretch{1}
    \setlength{\tabcolsep}{10pt}
  \begin{tabular}{c c ccc ccc ccc}
  \toprule
  \multirow{2}{*}{Method} & \multirow{2}{*}{Modality} & \multicolumn{3}{c}{Negative obstacle} & \multicolumn{3}{c}{Positive obstacle} & \multirow{2}{*}{mAcc} & \multirow{2}{*}{mIoU} & \multirow{2}{*}{mF1}  \\
\cmidrule(lr){3-5} \cmidrule(lr){6-8}
  & & Acc & IoU & F1 & Acc & IoU & F1 &  &  &  \\
   \midrule
  PotCrackSeg~\cite{feng2024tiv} & RGB+depth & 93.37 & 87.69 & 93.44 & 96.73 & 93.43 & 96.61 & 95.05 & 90.56 & 95.03  \\
  CLFT~\cite{gu2024tiv} & RGB+depth & 92.79 & 87.55 & 93.37 & 94.88 & 91.59 & 95.61 & 93.84 & 89.57 & 94.49  \\
  InconSeg~\cite{feng2023ral} & RGB+depth & 93.99 & 87.62 & 93.40 & 97.31 & 94.05 & 96.94 & 95.65 & 90.84 & 95.17  \\
  LuSeg(Ours) & RGB+depth & \textbf{94.12} & \textbf{88.35} & \textbf{93.81} & \textbf{98.91} & \textbf{97.23} & \textbf{98.59} & \textbf{96.52} & \textbf{92.79} & \textbf{96.20}\\
  \bottomrule
\end{tabular}
  \label{tab:LunarSeg}
\end{table*}

\begin{table*}[t]
  \caption{The Comparative Result(\%) on Different Scenes in the Testing Set of Our LunarSeg Dataset.}
  \centering
\renewcommand\arraystretch{1}
    \setlength{\tabcolsep}{5.6pt}
  \begin{tabular}{c c ccc ccc ccc ccc}
  \toprule
  \multirow{2}{*}{Method} & \multirow{2}{*}{Modality} & \multicolumn{3}{c}{Scene LR} & \multicolumn{3}{c}{Scene LF} & \multicolumn{3}{c}{Scene HR} & \multicolumn{3}{c}{Scene HF} \\
\cmidrule(lr){3-5} \cmidrule(lr){6-8}  \cmidrule(lr){9-11}  \cmidrule(lr){12-14}
  & & mAcc & mIoU & mF1 & mAcc & mIoU & mF1& mAcc & mIoU & mF1 & mAcc & mIoU & mF1  \\
   \midrule
  PotCrackSeg~\cite{feng2024tiv} & RGB+depth & 94.42 & 89.36 & 94.37 & 95.11 & 90.98 & 95.23 & 94.57 & 89.94 & 94.68 & 96.37 & 92.33 & 95.99 \\
  CLFT~\cite{gu2024tiv} & RGB+depth & 92.83 & 88.56 & 93.92 & 93.98 & 89.81 & 94.60 & 93.58 & 89.13 & 94.24 & 95.35 & 91.21 & 95.39 \\
  InconSeg~\cite{feng2023ral} & RGB+depth & 95.07 & 89.89 & 94.65 & 95.78 & 91.15 & 95.33 & 95.27 & 90.39 & 94.92 & 96.61 & 92.11 & 95.87 \\
  LuSeg(Ours) & RGB+depth & \textbf{96.21} & \textbf{92.13} & \textbf{95.86} & \textbf{96.44} & \textbf{92.72} & \textbf{96.15} & \textbf{96.27} & \textbf{92.73} & \textbf{96.17} & \textbf{97.25} & \textbf{93.99} & \textbf{96.86} \\
  \bottomrule
\end{tabular}
  \label{tab:LunarSeg_scenes}
\end{table*}

\begin{table*}[t]
  \caption{The Comparetive Result(\%) on the Testing Set of NPO Dataset.}
  \centering
\renewcommand\arraystretch{1}
    \setlength{\tabcolsep}{10pt}
  \begin{tabular}{c c ccc ccc ccc}
  \toprule
  \multirow{2}{*}{Method} & \multirow{2}{*}{Modality} & \multicolumn{3}{c}{Negative obstacle} & \multicolumn{3}{c}{Positive obstacle} & \multirow{2}{*}{mAcc} & \multirow{2}{*}{mIoU} & \multirow{2}{*}{mF1} \\
\cmidrule(lr){3-5} \cmidrule(lr){6-8}
  & & Acc & IoU & F1 & Acc & IoU & F1 &  &  &  \\
   \midrule
  PotCrackSeg~\cite{feng2024tiv} & RGB+depth & 89.01 & 58.79 & 74.04 & 97.16 & 91.70 & 95.67 & 93.09 & 75.25 & 84.86 \\
  CLFT~\cite{gu2024tiv} & RGB+depth & 78.02 & 60.87 & 75.68 & 93.34 & 86.89 & 92.98 & 85.68 & 73.88 & 84.33 \\
  InconSeg~\cite{feng2023ral} & RGB+depth & 83.43 & 74.51 & 85.39 & 95.87 & 93.04 & 96.38 & 89.65 & 83.74 & 90.89 \\
  LuSeg(Ours) & RGB+depth & \textbf{89.30} & \textbf{76.80} & \textbf{86.88} & \textbf{97.20} & \textbf{93.70} & \textbf{96.75} & \textbf{93.25} & \textbf{85.25} & \textbf{91.82} \\
  \bottomrule
\end{tabular}
  \label{tab:NPO}
\end{table*}

%

\subsection{Experiment Setups}
\subsubsection{Datasets} We conducted experiments on two datasets. First, our self-constructed LunarSeg dataset was collected using the LESS system for lunar obstacle segmentation. Additionally, we used the NPO dataset~\cite{feng2023ral}, a road scenario dataset designed for positive and negative obstacle segmentation, to validate the generalization ability of our method.

\textbf{LunarSeg Dataset}. Consistent with the processing methodology applied to the NPO dataset~\cite{feng2023ral}, the LunarSeg dataset is randomly divided into a training set (3,085 groups of RGB-D images), a validation set (1,028 groups of RGB-D images), and a test set (1,030 groups of RGB-D images). Throughout both the training and testing phases, the original resolution of 640×480 is preserved.

\textbf{NPO Dataset}. It was captured with an on-vehicle ZED stereo camera in Qingyuan, Fushun City, Liaoning Province, China. The dataset comprises 5,000 images collected from both urban and rural environments. It encompasses a wide range of road surface conditions (normal and abnormal) and various weather conditions (sunny, snowy, and cloudy), thereby providing a comprehensive resource for obstacle segmentation research across diverse real-world driving scenarios. Following the methodology established in InconSeg~\cite{feng2023ral}, we divide the NPO dataset into three subsets: 2,500 RGB-D image pairs for training, 1,250 for validation and 1,250 for testing. The image resolution is reduced to 512 × 288 during both the training and testing phases.

\subsubsection{Implementation Details} We implemented our proposed LuSeg model using PyTorch. The network was trained and evaluated on a workstation equipped with an NVIDIA RTX 4090 GPU. Following InconSeg~\cite{feng2023ral}, we initialized the parameters of the first four encoder stages with pre-trained weights from the ResNet~\cite{he2016cvpr} model provided by PyTorch. For optimization, we utilized the stochastic gradient descent (SGD) optimizer, setting the initial learning rate, momentum, and decay factor to 0.01, 0.90, and 0.95, respectively.


\subsection{Obstacles Segmentation Performance}
We conducted extensive comparative analyses of our proposed LuSeg model against three baseline obstacle segmentation methods: PotCrackSeg~\cite{feng2024tiv}, CLFT~\cite{gu2024tiv}, and InconSeg~\cite{feng2023ral}. PotCrackSeg adopts a dual-modal semantic feature complementary fusion strategy, CLFT employs a progressive assembly multi-level fusion method, while InconSeg incorporates a residual structure within its dual decoders to enhance feature fusion. We use Acc, F1, and IoU metrics to evaluate the segmentation performance for negative and positive obstacles.



\begin{table}[t]
  \caption{Ablation Study on Different Datasets.}
   \centering
\setlength{\tabcolsep}{4pt}
\begin{tabular}{c c c c | c c c | c}
  \toprule
 Dataset &RGB & Depth & CDFM & mAcc & mIoU & mF1 &\makecell{Runtime \\ (ms)} \\
  \midrule
    \multirow{4}{*}{LunarSeg} &  & \checkmark & & 95.61 & 90.92 & 95.14 &10.46 \\
  & \checkmark & & & 96.27 & 91.79 & 95.68 & 14.41\\
  & \checkmark & \checkmark & & 95.89 & 92.24 & 95.88 & 17.47 \\
  & \checkmark & \checkmark & \checkmark & \textbf{96.52} & \textbf{92.79} & \textbf{96.20} & 17.45 \\
  \midrule
  \multirow{4}{*}{NPO} &  & \checkmark & & 87.99 & 69.49 & 80.95 & 7.69\\
  & \checkmark & & & 93.17 & 83.99 & 91.03 & 10.06\\
  & \checkmark & \checkmark & & 93.06 & 84.56 & 91.41 & 12.94 \\
  & \checkmark & \checkmark & \checkmark & \textbf{93.25} & \textbf{85.25} & \textbf{91.82} & 12.93 \\
  \bottomrule
  
\end{tabular}

  \label{tab:ablation}
\end{table}

\begin{figure*}[t]
  \centering
  \includegraphics[width=1\linewidth]{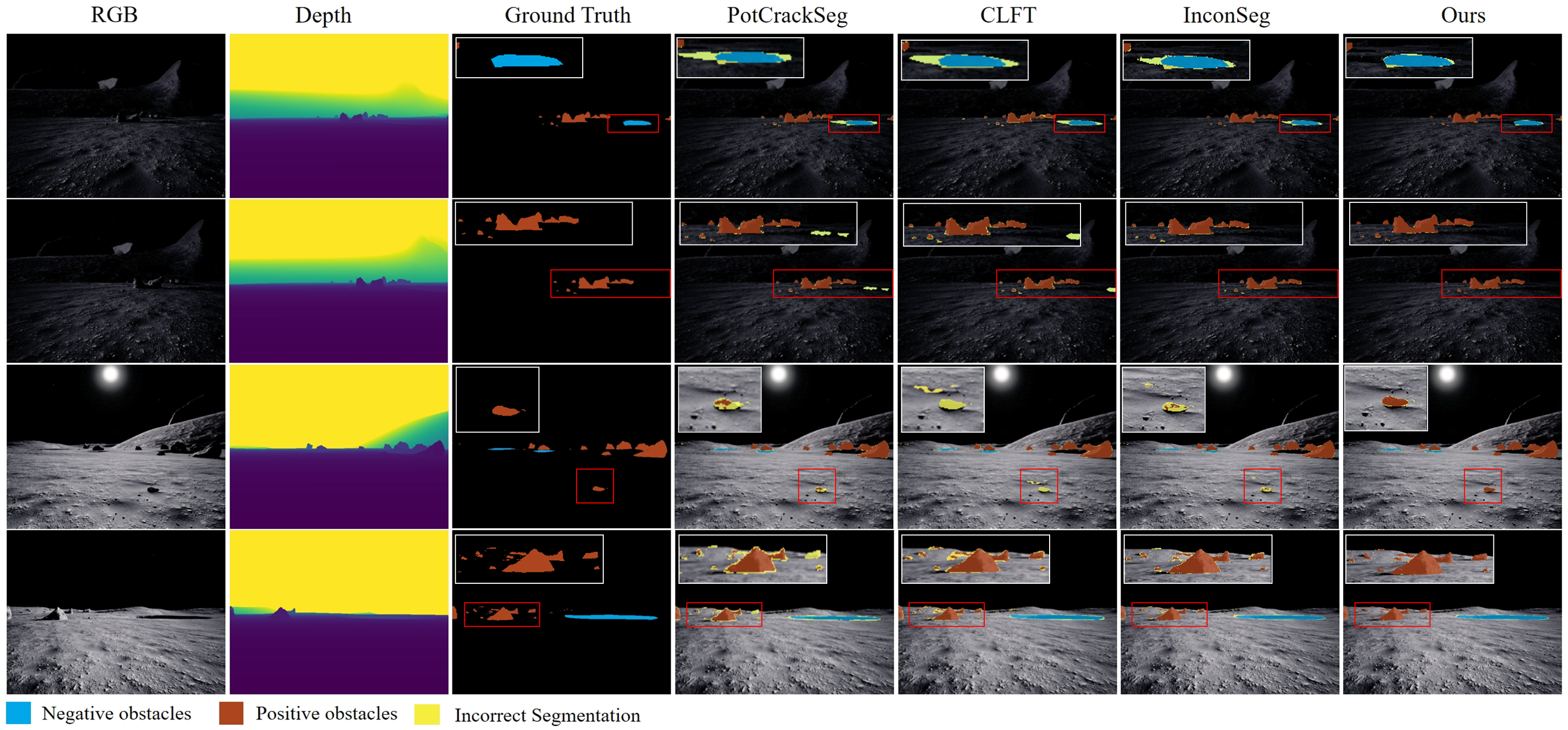}
  \caption{Sample qualitative results of our LunarSeg dataset. The white box in the image is a magnified view of the red box.}
  \label{fig:vis}
\end{figure*}

\subsubsection{The Overall Results on our LunarSeg dataset} We assessed the performance of our method for obstacle segmentation in simulated lunar scenarios using the LunarSeg dataset. These scenarios are characterized by their unstructured nature and poor lighting conditions. As shown in \cref{tab:LunarSeg}, our method consistently surpassed all baseline approaches across multiple metrics, underscoring its robustness and reliability. Additionally, we also evaluated our method using images from different scenarios, with the results presented in \cref{tab:LunarSeg_scenes}. Our method exhibits remarkable performance across a range of scenarios, indicating its capability to effectively tackle the task of lunar obstacle segmentation.

\subsubsection{The Overall Results on the NPO dataset} We also evaluated the performance of our method for obstacle segmentation in real-world road scenarios using the NPO dataset to validate the generalization ability of our method. \cref{tab:NPO} presents the results of all baseline methods trained and tested on the NPO dataset. It can be observed that our approach significantly outperforms the compared approaches across all evaluation metrics. Compared to the suboptimal method, InconSeg, our LuSeg improves the mIoU from 83.74\% to 85.25\%. This implies that although these handcrafted feature fusion modules can integrate semantic information from different modalities, their alignment may not be optimal. In contrast, our approach leverages a contrastive learning mechanism to significantly enhance the semantic consistency between features from the two modalities, thereby demonstrating superior segmentation performance.


\subsubsection{The  Qualitative Results}
\begin{figure}[t]
  \centering
  \includegraphics[width=0.99\linewidth]{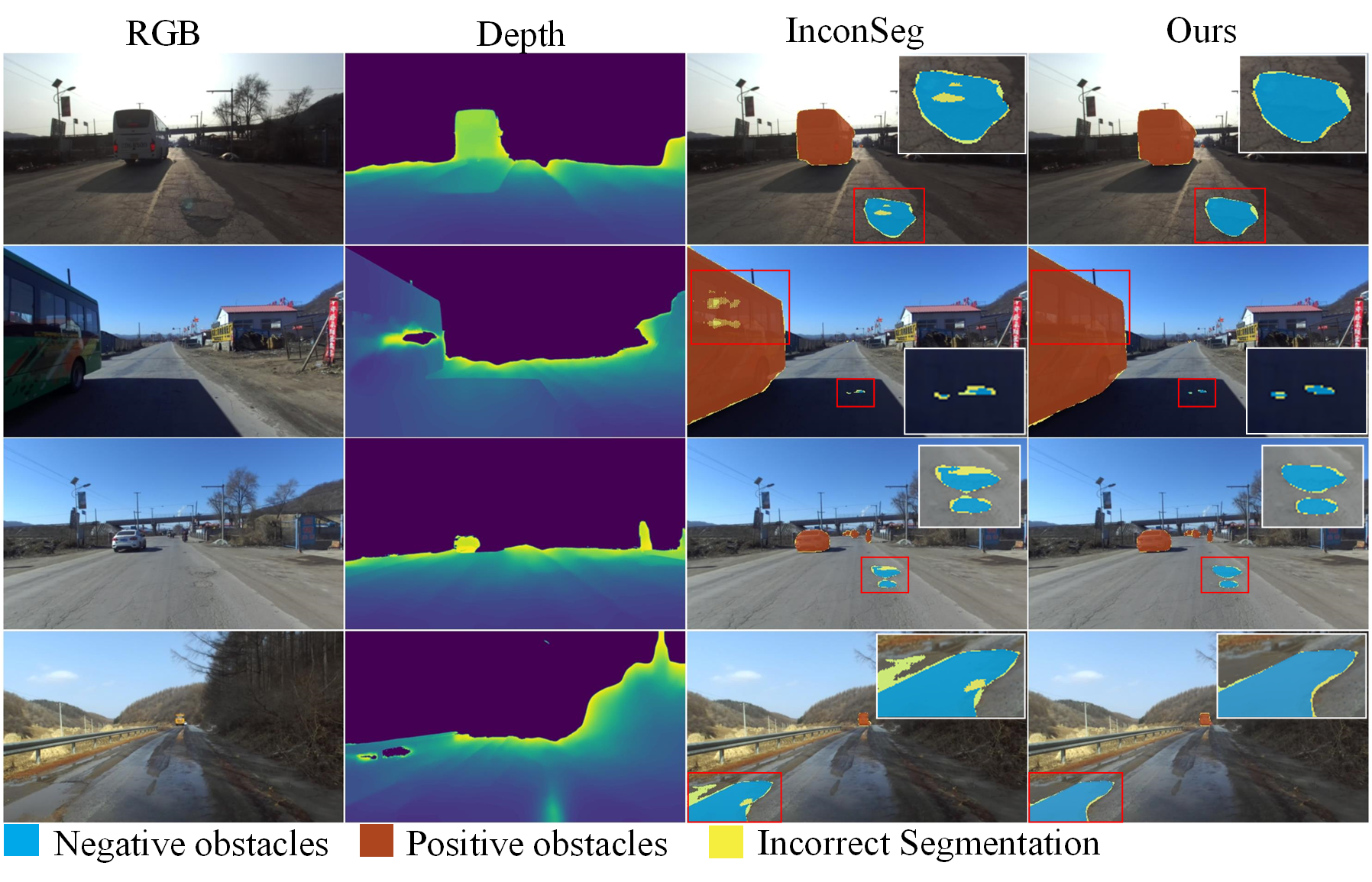}
  \caption{Sample qualitative results of NPO dataset. The white box in the image is a magnified view of the red box.}
  \label{fig:npo_vis}
\end{figure}

~\cref{fig:vis} illustrates the qualitative results of samples from different scenarios in our LunarSeg dataset, demonstrating that our method consistently outperforms the baseline methods in segmentation results, particularly showing outstanding performance in segmenting small targets (such as small lunar rocks). We additionally present several qualitative results from the NPO dataset, as shown in ~\cref{fig:npo_vis}. These results indicate that our contrastive learning-driven fusion method significantly enhances the model's discriminative ability, particularly demonstrating superior performance in scenarios with subtle or ambiguous boundaries.

\begin{table}[t]
  \caption{Runtime Performance on the LunarSeg Dataset.}
   \centering
\setlength{\tabcolsep}{35pt}
\begin{tabular}{c | c }
  \toprule
  Method  &Runtime (ms) \\
  \midrule
  PotCrackSeg~\cite{feng2024tiv} & 18.83\\
  CLFT~\cite{gu2024tiv}  & 12.25\\
InconSeg~\cite{feng2023ral}  & 23.87 \\
LuSeg(Ours) & 17.45 \\
  \bottomrule
\end{tabular}

  \label{tab:runtime}
\end{table}

\subsection{Ablation Study} We conducted a series of ablation experiments on different datasets to assess the effectiveness of our proposed CDFM module. The experiments were designed in a sequential manner: first, we tested a single depth modality, followed by a single RGB modality, then a dual-modality fusion, and finally, a dual modality fusion incorporating the CDFM module. The results, presented in \cref{tab:ablation}, demonstrate that the CDFM module significantly enhances performance. Additionally, we performed inference time evaluations, with the results displayed in the final column of \cref{tab:ablation}, showing that our CDFM module preserves model performance without introducing any additional computational overhead.

\subsection{Runtime} Real-time detection of positive and negative obstacles is a crucial prerequisite for ensuring the safe autonomous driving of lunar rovers. We evaluated the runtime requirements of our method using our LunarSeg dataset.  \cref{tab:runtime} reports the runtime performance of LuSeg compared to baseline methods. To ensure fair comparisons, all experiments were conducted on a computer equipped with an Intel i7-13700KF CPU and an NVIDIA RTX4090 GPU. As depicted in \cref{tab:runtime}, our approach not only achieves superior segmentation performance but also operates with a significantly faster inference speed of 57 Hz, compared to methods like PotCrackSeg~\cite{feng2024tiv} and InconSeg~\cite{feng2023ral}. It is noteworthy that this high speed is markedly faster than the typical sampling rate of mainstream cameras, which generally operate at around 30 Hz.


\section{Conclusion}
\label{sec:conclusion}

In this paper, we proposed a Lunar Exploration Simulation System (LESS) which is compatible with the Robot Operating System (ROS) and different robotic sensors, enabling the customization of scenes and environments tailored to specific tasks. Furthermore, we released a novel lunar segmentation dataset for positive and negative obstacles, called LunarSeg. Additionally, we proposed a new accurate and efficient Negative and Positive Obstacles segmentation method, named LuSeg, which effectively maintains the semantic consistency of multimodal features via our proposed Contrast-Driven Fusion module. We evaluated our LuSeg on the LunarSeg dataset and the additional public real-world NPO road obstacle dataset. The experimental results demonstrate that the proposed method achieves state-of-the-art performance on both datasets while delivering a faster inference speed of approximately 57 Hz. The LESS system, LunarSeg dataset, and LuSeg network implementation have been open-sourced to support and advance future research.



\bibliographystyle{ieeetr}

\bibliography{paper}

\begin{thebibliography}{10}

\bibitem{frank2016AI}
J.~D. Frank, K.~McGuire, H.~R. Moses, and J.~Stephenson, ``Developing decision aids to enable human spaceflight autonomy,'' {\em AI Magazine}, vol.~37, no.~4, pp.~46--54, 2016.

\bibitem{crues2022dles}
E.~Z. Crues, S.~J. Lawrence, P.~Bielski, A.~B. Jacobs, J.~Schlueter, A.~Jagge, C.~Foreman, C.~Raymond, and N.~Davis, ``Digital lunar exploration sites (dles),'' in {\em 2022 IEEE Aerospace Conference (AERO)}, pp.~1--13, IEEE, 2022.

\bibitem{bingham2023dust}
L.~Bingham, J.~Kincaid, B.~Weno, N.~Davis, E.~Paddock, and C.~Foreman, ``Digital lunar exploration sites unreal simulation tool (dust),'' in {\em 2023 IEEE Aerospace Conference}, pp.~1--12, IEEE, 2023.

\bibitem{Allan2019PlanetaryRS}
M.~Allan, U.~Y. Wong, P.~M. Furlong, A.~Rogg, S.~T. McMichael, T.~M. Welsh, I.~Chen, S.~C. Peters, B.~P. Gerkey, M.~Quigley, M.~H. Shirley, M.~C. Deans, H.~N. Cannon, and T.~Fong, ``Planetary rover simulation for lunar exploration missions,'' {\em 2019 IEEE Aerospace Conference}, pp.~1--19, 2019.

\bibitem{martinez2023multi}
B.~Martinez Rocamora~Jr, C.~Kilic, C.~Tatsch, G.~A. Pereira, and J.~N. Gross, ``Multi-robot cooperation for lunar in-situ resource utilization,'' {\em Frontiers in Robotics and AI}, vol.~10, p.~1149080, 2023.

\bibitem{orsula2022iros}
A.~Orsula, S.~B{\o}gh, M.~Olivares-Mendez, and C.~Martinez, ``Learning to grasp on the moon from 3d octree observations with deep reinforcement learning,'' in {\em Proc.~of the IEEE/RSJ Intl.~Conf.~on Intelligent Robots and Systems (IROS)}, pp.~4112--4119, IEEE, 2022.

\bibitem{shah2018airsim}
S.~Shah, D.~Dey, C.~Lovett, and A.~Kapoor, ``Airsim: High-fidelity visual and physical simulation for autonomous vehicles,'' in {\em Field and Service Robotics: Results of the 11th International Conference}, pp.~621--635, Springer, 2018.

\bibitem{pessia2020}
R.~Pessia {\em et~al.}, ``https://www.kaggle.com/datasets/romainpessia/artificial-lunar-rocky-landscape-dataset.'' Online, 2020.

\bibitem{boerdijk2023resyris}
W.~Boerdijk, M.~G. M{\"u}ller, M.~Durner, and R.~Triebel, ``Resyris-a real-synthetic rock instance segmentation dataset for training and benchmarking,'' in {\em 2023 IEEE Aerospace Conference}, pp.~1--9, IEEE, 2023.

\bibitem{muhammad2022tits}
K.~Muhammad, T.~Hussain, H.~Ullah, J.~Del~Ser, M.~Rezaei, N.~Kumar, M.~Hijji, P.~Bellavista, and V.~H.~C. de~Albuquerque, ``Vision-based semantic segmentation in scene understanding for autonomous driving: Recent achievements, challenges, and outlooks,'' {\em IEEE Trans.~on Intelligent Transportation Systems (T-ITS)}, vol.~23, no.~12, pp.~22694--22715, 2022.

\bibitem{ma2022computer}
N.~Ma, J.~Fan, W.~Wang, J.~Wu, Y.~Jiang, L.~Xie, and R.~Fan, ``Computer vision for road imaging and pothole detection: a state-of-the-art review of systems and algorithms,'' {\em Transportation safety and Environment}, vol.~4, no.~4, p.~tdac026, 2022.

\bibitem{seichter2021icra}
D.~Seichter, M.~K{\"o}hler, B.~Lewandowski, T.~Wengefeld, and H.-M. Gross, ``Efficient rgb-d semantic segmentation for indoor scene analysis,'' in {\em Proc.~of the IEEE Intl.~Conf.~on Robotics \& Automation (ICRA)}, pp.~13525--13531, IEEE, 2021.

\bibitem{zhou2023tase}
W.~Zhou, Y.~Xiao, W.~Yan, and L.~Yu, ``Cmpffnet: Cross-modal and progressive feature fusion network for rgb-d indoor scene semantic segmentation,'' {\em IEEE Transactions on Automation Science and Engineering}, 2023.

\bibitem{couprie2013iclr}
C.~Couprie, C.~Farabet, L.~Najman, and Y.~LeCun, ``Indoor semantic segmentation using depth information: 1st international conference on learning representations, iclr 2013,'' in {\em 1st International Conference on Learning Representations, ICLR 2013}, 2013.

\bibitem{valada2017deep}
A.~Valada, G.~L. Oliveira, T.~Brox, and W.~Burgard, ``Deep multispectral semantic scene understanding of forested environments using multimodal fusion,'' in {\em 2016 International Symposium on Experimental Robotics}, pp.~465--477, Springer, 2017.

\bibitem{cheng2017cvpr}
Y.~Cheng, R.~Cai, Z.~Li, X.~Zhao, and K.~Huang, ``Locality-sensitive deconvolution networks with gated fusion for rgb-d indoor semantic segmentation,'' in {\em Proc.~of the IEEE/CVF Conf.~on Computer Vision and Pattern Recognition (CVPR)}, pp.~3029--3037, 2017.

\bibitem{zhou2023tiv}
W.~Zhou, S.~Dong, M.~Fang, and L.~Yu, ``Cacfnet: Cross-modal attention cascaded fusion network for rgb-t urban scene parsing,'' {\em IEEE Trans.~on Intelligent Vehicles}, 2023.

\bibitem{du2024cvpr}
S.~Du, W.~Wang, R.~Guo, R.~Wang, and S.~Tang, ``Asymformer: Asymmetrical cross-modal representation learning for mobile platform real-time rgb-d semantic segmentation,'' in {\em Proc.~of the IEEE/CVF Conf.~on Computer Vision and Pattern Recognition (CVPR)}, pp.~7608--7615, 2024.

\bibitem{hoyer2022cvpr}
L.~Hoyer, D.~Dai, and L.~Van~Gool, ``Daformer: Improving network architectures and training strategies for domain-adaptive semantic segmentation,'' in {\em Proc.~of the IEEE/CVF Conf.~on Computer Vision and Pattern Recognition (CVPR)}, pp.~9924--9935, 2022.

\bibitem{zhang2022laanet}
X.~Zhang, B.~Du, Z.~Wu, and T.~Wan, ``Laanet: lightweight attention-guided asymmetric network for real-time semantic segmentation,'' {\em Neural Computing and Applications}, vol.~34, no.~5, pp.~3573--3587, 2022.

\bibitem{xu2023cvpr}
J.~Xu, Z.~Xiong, and S.~P. Bhattacharyya, ``Pidnet: A real-time semantic segmentation network inspired by pid controllers,'' in {\em Proc.~of the IEEE/CVF Conf.~on Computer Vision and Pattern Recognition (CVPR)}, pp.~19529--19539, 2023.

\bibitem{qian2021tits}
Y.~Qian, L.~Deng, T.~Li, C.~Wang, and M.~Yang, ``Gated-residual block for semantic segmentation using rgb-d data,'' {\em IEEE Trans.~on Intelligent Transportation Systems (T-ITS)}, vol.~23, no.~8, pp.~11836--11844, 2021.

\bibitem{zhou2023if}
W.~Zhou, Y.~Yue, M.~Fang, X.~Qian, R.~Yang, and L.~Yu, ``Bcinet: Bilateral cross-modal interaction network for indoor scene understanding in rgb-d images,'' {\em Information Fusion}, vol.~94, pp.~32--42, 2023.

\bibitem{zhang2023sj}
Y.~Zhang, C.~Xiong, J.~Liu, X.~Ye, and G.~Sun, ``Spatial-information guided adaptive context-aware network for efficient rgb-d semantic segmentation,'' {\em IEEE Sensors Journal}, 2023.

\bibitem{muller2021iros}
M.~G. M{\"u}ller, M.~Durner, A.~Gawel, W.~St{\"u}rzl, R.~Triebel, and R.~Siegwart, ``A photorealistic terrain simulation pipeline for unstructured outdoor environments,'' in {\em Proc.~of the IEEE/RSJ Intl.~Conf.~on Intelligent Robots and Systems (IROS)}, pp.~9765--9772, IEEE, 2021.

\bibitem{swan2021cvpr}
R.~M. Swan, D.~Atha, H.~A. Leopold, M.~Gildner, S.~Oij, C.~Chiu, and M.~Ono, ``Ai4mars: A dataset for terrain-aware autonomous driving on mars,'' in {\em Proc.~of the IEEE/CVF Conf.~on Computer Vision and Pattern Recognition (CVPR)}, pp.~1982--1991, 2021.

\bibitem{he2016cvpr}
K.~He, X.~Zhang, S.~Ren, and J.~Sun, ``{Deep Residual Learning for Image Recognition},'' in {\em Proc.~of the IEEE Conf.~on Computer Vision and Pattern Recognition (CVPR)}, 2016.

\bibitem{feng2023ral}
Z.~Feng, Y.~Guo, D.~Navarro-Alarcon, Y.~Lyu, and Y.~Sun, ``Inconseg: Residual-guided fusion with inconsistent multi-modal data for negative and positive road obstacles segmentation,'' {\em IEEE Robotics and Automation Letters (RA-L)}, vol.~8, no.~8, pp.~4871--4878, 2023.

\bibitem{chen2020icml}
T.~Chen, S.~Kornblith, M.~Norouzi, and G.~Hinton, ``{A Simple Framework for Contrastive Learning of Visual Representations},'' in {\em Proc.~of the Intl.~Conf.~on Machine Learning (ICML)}, 2020.

\bibitem{berman2018cvpr}
M.~Berman, A.~R. Triki, and M.~B. Blaschko, ``{The Lovász-Softmax Loss: A Tractable Surrogate for the Optimization of the Intersection-Over-Union Measure in Neural Networks},'' in {\em Proc.~of the IEEE/CVF Conf.~on Computer Vision and Pattern Recognition (CVPR)}, 2018.

\bibitem{feng2024tiv}
Z.~Feng, Y.~Guo, and Y.~Sun, ``Segmentation of road negative obstacles based on dual semantic-feature complementary fusion for autonomous driving,'' {\em IEEE Trans.~on Intelligent Vehicles}, 2024.

\bibitem{gu2024tiv}
J.~Gu, M.~Bellone, T.~Pivo{\v{n}}ka, and R.~Sell, ``Clft: Camera-lidar fusion transformer for semantic segmentation in autonomous driving,'' {\em IEEE Trans.~on Intelligent Vehicles}, 2024.

\end{thebibliography}

\end{document}